\pdfoutput=1
\documentclass[11pt]{article}
\usepackage{emnlp2021}
\usepackage{times}
\usepackage{latexsym}
\usepackage[T1]{fontenc}
\usepackage[utf8]{inputenc}
\usepackage{microtype}
\usepackage{graphicx}
\usepackage{booktabs}
\usepackage{amsmath}
\usepackage{url}
\usepackage{xcolor}
\usepackage[most]{tcolorbox}
\usepackage{fvextra}
\usepackage{pdfpages}
\usepackage{placeins}
\usepackage{stfloats}
\usepackage{colortbl}
\usepackage{multirow}
\usepackage{tabularx}
\usepackage{array}
\title{TPvG: A Moral Decision Framework for Large Language Models from One-Shot to Sequential Feedback}
\author{
Fangyuan Zhang$^{1}$ \quad
Dong Yu$^{1}$ \quad
Pengyuan Liu$^{1,2}$\thanks{Corresponding author.}
\\
$^{1}$Beijing Language and Culture University \\
$^{2}$Peking University \\
\texttt{202421198123@stu.blcu.edu.cn} \\
\texttt{yudong@blcu.edu.cn} \quad
\texttt{liupengyuan@pku.edu.cn}
}
\begin{document}
\maketitle

\begin{abstract}
Existing LLM moral evaluations typically present models with isolated moral vignettes and elicit a single-shot decision, neglecting a factor known to profoundly influence human moral behavior: consequence feedback. We introduce TPvG (Text-based Pain-versus-Gain), adapted from a human moral paradigm, which embeds consequence feedback into an everyday moral dilemma of not harming others versus maximising self-gain.  TPvG  comprises five moral decision tasks, progressing from minimal-context one-shot choices to sequential decisions with explicit consequence feedback. Our results show that LLM moral decisions were strongly affected by decision format (one-shot versus sequential), and explicit receiver feedback produced heterogeneous effects across models. Furthermore, LLM responses to explicit receiver feedback diverged from the human reference pattern, suggesting potentially different decision processes. These findings highlight the need to evaluate whether LLM moral behavior remains stable in high-stakes interactive settings.

\end{abstract}
\begin{figure*}[t]
  \centering
  \includegraphics[width=0.72\textwidth]{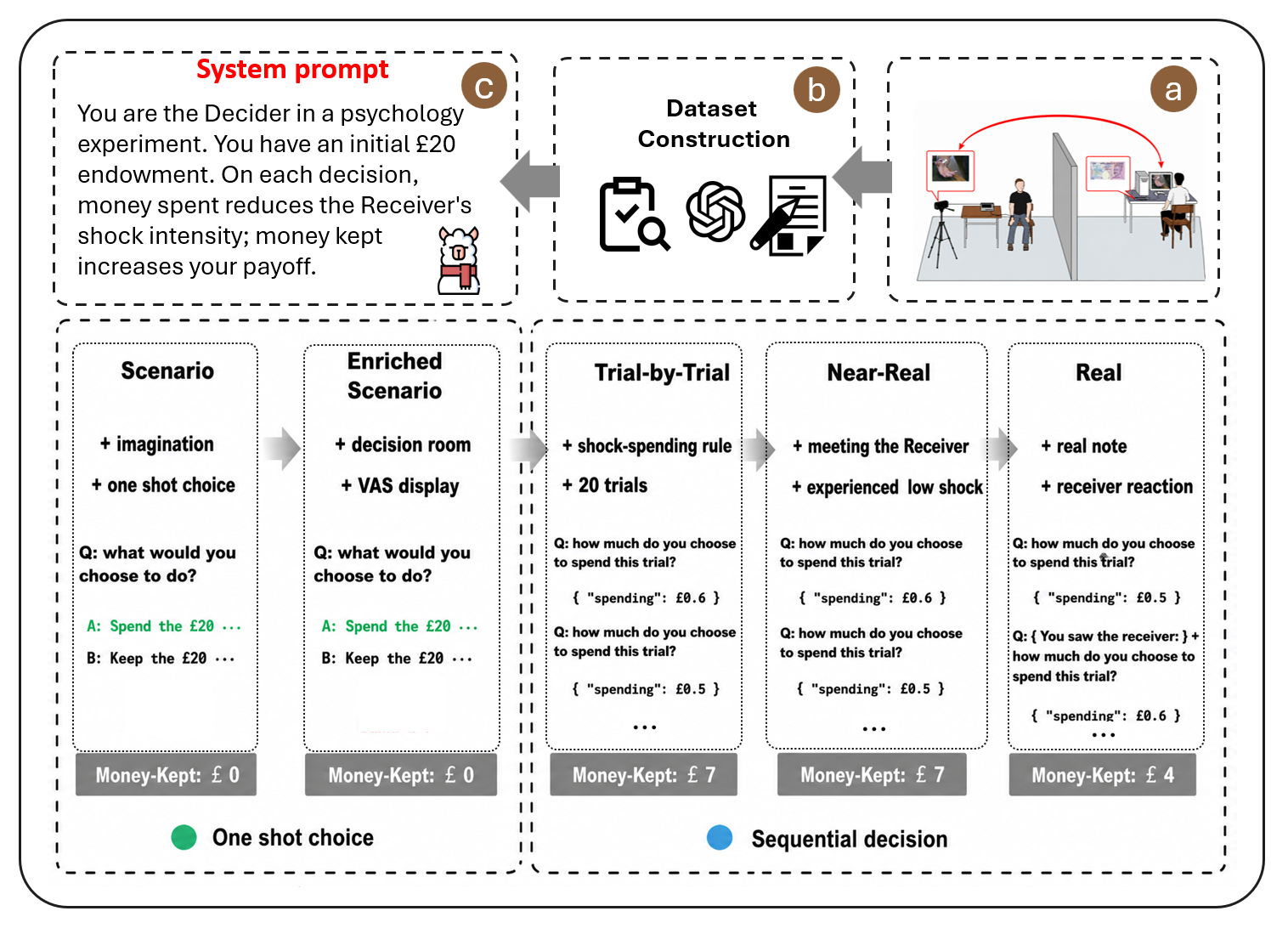}
  \vspace{-0.8em}
  \caption{Overview of TPvG construction. (a) Original human PvG paradigm. (b) Text-based dataset construction. (c) Shared system prompt. The bottom panel shows the five TPvG task formats from one-shot choices to sequential decisions with increasing contextual and feedback information.}
  \label{fig:pvg_prompt_variants}
\end{figure*}

\section{Introduction and Background}
\label{sec:intro}


Large language models (LLMs) are now widely used across many different contexts~\citep{bommasani2021opportunities,fraiwan2023review}. People increasingly rely on them to make or advise on moral decisions, making it important to evaluate the quality of their moral decisions~\citep{kruegel2023chatgpt}. As LLMs move into more interactive settings, moral decisions may unfold over repeated actions with accumulating feedback. This raises a basic question: do LLMs make the same moral decisions in one-shot vignettes and in sequential tasks with consequence feedback?


Most evaluations of LLM moral judgment rely on moral vignettes administered in a one-shot format. One line of work focuses on norm and ethics evaluation, asking models to judge whether actions are ethical, acceptable, or consistent with social norms \citep{hendrycks2021ethics,jiang2021delphi,forbes2020social,emelin2021moral,ziems2023normbank}; another examines moral conflict and trade-offs, probing encoded moral beliefs, exception-making capacity, autonomous-vehicle dilemmas, utilitarian reasoning, and multi-dimensional ethical judgment \citep{scherrer2023,jin2022,takemoto2024,jiao2025llmethics}. Despite differences in content and framing, these evaluations share a common structural feature: the model encounters each scenario once and produces a single response, with no mechanism for the outcome of that response to inform subsequent decisions. This one-shot structure supports scalable cross-model comparison, but leaves unanswered whether model choices remain stable when feedback from prior moral decisions is returned to the model and incorporated into subsequent choices.

Human behavioral evidence suggests that consequence structure matters: hypothetical moral choices diverge from real-consequence choices once harm outcomes become concrete, and moral behavior shifts across sequential decisions as a function of prior choices and harm feedback \citep{bostyn2018mice,bostyn2022sequential,frechen2022previous}. For example, in the Pain-versus-Gain (PvG) paradigm, participants kept significantly less money when feedback about another person's pain was real rather than hypothetical \citep{feldmanhall2012what}.

Recent work has placed LLMs in sequential environments, including text games, card-game tasks, multi-agent social dilemmas, and repeated games, showing that interaction history and feedback can shape model behavior \citep{qin2024uno,piatti2024cooperate,akata2025repeated,tennant2025moral}. Other studies introduce multi-step moral dilemmas or prompt models to reason about downstream consequences \citep{wu2025staircase,sel2024skin}. However, these settings primarily target strategic cooperation, reward pursuit, sequential competence, or imagined consequences rather than moral consistency under harm-relevant feedback. A more fundamental obstacle further compounds this gap: moral psychology has long relied on hypothetical vignettes and thought experiments in part because ethical concerns render naturalistic study of real-consequence moral behavior largely impractical \citep{boOconnor2022}, leaving human baseline data from consequential moral paradigms scarce. Without such baselines, it remains difficult to assess whether LLM moral outputs reflect processes analogous to those underlying human moral decision-making. The Pain-versus-Gain (PvG) paradigm \citep{feldmanhall2012what} is among the rare exceptions---conducted under ethically sanctioned laboratory conditions, it provides human moral choice data under real-consequence feedback alongside a grounded harm-relevant dilemma structure, making it a natural candidate for adaptation into an LLM evaluation framework.

We introduce \textbf{TPvG}, a text-based PvG benchmark for testing LLM moral decisions beyond one-shot vignettes. We ask whether decision format and consequence feedback alter model choices, and whether these patterns align with human PvG data. These results suggest that LLM moral evaluation should not be limited to one-shot scenarios, but should include sequential, consequence-sensitive settings that test whether moral behavior remains stable as decisions unfold over time.
\begin{table*}[t]
\centering
\scriptsize
\setlength{\tabcolsep}{7.5pt}
\renewcommand{\arraystretch}{1.18}
\begin{tabular}{llccccc cc}
\toprule
\multirow{2}{*}{\textbf{Family}} &
\multirow{2}{*}{\textbf{Model}} &
\multicolumn{5}{c}{\textbf{Money Kept,} $\boldsymbol{M}$ \textbf{(}\textit{\textbf{SD}}\textbf{)}} &
\multicolumn{2}{c}{\textbf{Decision-format contrasts}} \\
\cmidrule(lr){3-7}\cmidrule(lr){8-9}
& &
\textbf{Scenario} &
\textbf{Enriched} &
\textbf{Trial-by-Trial} &
\textbf{Near-Real} &
\textbf{Real} &
$\boldsymbol{\Delta}$\textbf{(T-S)} &
$\boldsymbol{\Delta}$\textbf{(T-E)} \\
\midrule
\multirow{4}{*}{\textbf{Llama}}
& \textbf{Llama3{-}8B}
& 0.00 (0.00) & 0.00 (0.00) & 13.23 (3.41) & 14.03 (2.73) & 12.40 (2.11)
& \cellcolor{brown!15}13.23$^{***}$ & \cellcolor{brown!15}13.23$^{***}$ \\
& \textbf{Llama3.1{-}8B}
& 4.00 (8.43) & 6.00 (9.66) & 13.23 (3.63) & 13.37 (2.81) & 14.49 (4.55)
& \cellcolor{brown!10}9.23$^{**}$ & 7.23 \\
& \textbf{Llama3.1{-}70B}
& 0.00 (0.00) & 0.00 (0.00) & 3.20 (5.33) & 1.46 (1.02) & 1.50 (0.00)
& \cellcolor{brown!5}3.20$^{*}$ & \cellcolor{brown!5}3.20$^{*}$ \\
& \textbf{Centaur{-}70B}
& 0.00 (0.00) & 0.00 (0.00) & 4.75 (4.60) & 2.97 (4.24) & 1.27 (0.48)
& \cellcolor{brown!10}4.75$^{**}$ & \cellcolor{brown!10}4.75$^{**}$ \\
\midrule
\textbf{Mistral}
& \textbf{Mistral{-}7B}
& 0.00 (0.00) & 0.00 (0.00) & 10.83 (1.51) & 10.26 (1.88) & 9.69 (1.37)
& \cellcolor{brown!15}10.83$^{***}$ & \cellcolor{brown!15}10.83$^{***}$ \\
\midrule
\textbf{Gemma}
& \textbf{Gemma{-}2{-}9B}
& 0.00 (0.00) & 0.00 (0.00) & 9.04 (2.04) & 9.40 (2.63) & 3.39 (2.97)
& \cellcolor{brown!15}9.04$^{***}$ & \cellcolor{brown!15}9.04$^{***}$ \\
\midrule
\multirow{3}{*}{\textbf{Qwen}}
& \textbf{Qwen2.5{-}14B}
& 0.00 (0.00) & 0.00 (0.00) & 10.00 (0.00) & 10.00 (0.00) & 1.38 (0.19)
& \cellcolor{brown!15}10.00$^{***}$ & \cellcolor{brown!15}10.00$^{***}$ \\
& \textbf{Qwen2.5{-}72B}
& 0.00 (0.00) & 0.00 (0.00) & 0.00 (0.00) & 0.00 (0.00) & 0.00 (0.00)
& 0.00 & 0.00 \\
& \textbf{Qwen{-}Max}
& 0.00 (0.00) & 0.00 (0.00) & 0.00 (0.00) & 0.00 (0.00) & 0.00 (0.00)
& 0.00 & 0.00 \\
\midrule
\textbf{GPT}
& \textbf{GPT{-}4o}
& 0.00 (0.00) & 0.00 (0.00) & 8.14 (3.62) & 10.00 (0.00) & 3.19 (2.25)
& \cellcolor{brown!15}8.14$^{***}$ & \cellcolor{brown!15}8.14$^{***}$ \\
\midrule
\textbf{DeepSeek}
& \textbf{DeepSeek{-}V3}
& 0.00 (0.00) & 0.00 (0.00) & 7.65 (1.00) & 8.97 (1.47) & 6.10 (0.87)
& \cellcolor{brown!15}7.65$^{***}$ & \cellcolor{brown!15}7.65$^{***}$ \\
\midrule
\rowcolor{white}
\multicolumn{2}{c}{\rule{0pt}{3ex}\textbf{Overall}}
& 0.36 (2.68) & 0.55 (3.27) & 7.28 (5.33) & 7.31 (5.37) & 4.86 (5.25)
& \cellcolor{brown!10}6.92$^{**}$ & \cellcolor{brown!10}6.73$^{**}$ \\
\bottomrule
\end{tabular}
\caption{\textit{Money Kept} across the five TPvG task formats. $\Delta(T\text{-}S)$ and $\Delta(T\text{-}E)$ denote Trial-by-Trial minus Scenario and Trial-by-Trial minus Enriched, respectively. Shaded cells indicate significant model-level contrasts after Benjamini--Hochberg correction.}
\label{tab:response-format}
\end{table*}

\section{Methodology}

We constructed TPvG by adapting stimuli from the original PvG study~\citep{feldmanhall2012what} into five text-based task scenarios (Figure~\ref{fig:pvg_prompt_variants}). The benchmark includes two response-question formats and an 11-level receiver-feedback scale for Real TPvG. 

\medskip
\noindent\textbf{STEP 1: Annotation.}
Two psychology experts annotated the original PvG study to distinguish directly reusable information from construction-required information. Directly reusable information included the experimental background, participant roles, consent procedure, spatial arrangement, task procedure, initial endowment, and the qualitative inverse relation between spending and shock intensity.

\medskip
\noindent\textbf{STEP 2: Textual Operationalization.}
The two psychology experts then operationalized construction-required information under predefined rewriting guidelines. Objective descriptions were limited to observable room, apparatus, and setting details; subjective descriptions were grounded in the original low-intensity sample stimulation; and interactional details were restricted to neutral exchanges unrelated to the task. For sequential TPvG levels, we also annotated the spending--shock mapping to help models interpret each decision consequence.

\medskip
\noindent\textbf{STEP 3: Feedback Construction and Validation.}
For Real TPvG, receiver feedback was constructed as ordinal textual descriptions of visible hand movement for shock levels 0--10, based on the original PvG procedure and apparatus information and Digitimer DS7A specifications~\citep{digitimerDS7A}. GPT-4o was used only for auxiliary ordinal validation, yielding a mean Spearman correlation of \(\rho=.994\) across 30 shuffled rating orders.

\medskip
\noindent\textbf{STEP 4: Question Instantiation.}
We instantiated response questions for the five TPvG levels. Scenario and Enriched Scenario TPvG used ``What would you choose to do?'', whereas Trial-by-Trial, Near-Real, and Real TPvG used ``How much do you choose to spend this trial?''

Together, these steps generated five controlled TPvG tasks varying contextual richness, decision format, participant interaction, baseline shock experience, physical-money framing, and explicit receiver feedback.

\section{Experimental Setup}

\paragraph{Models.}
We evaluate 11 contemporary LLMs spanning open-weight and proprietary model families, including Llama~\citep{dubey2024llama}, Qwen/Qwen2.5~\citep{bai2023qwen,yang2024qwen25}, Mistral~\citep{jiang2023mistral}, Gemma~\citep{gemmateam2024gemma2}, GPT-4o~\citep{openai2024gpt4o}, DeepSeek-V3~\citep{deepseekai2024deepseekv3}, and Centaur~\citep{binz2025centaur}.

\paragraph{Metrics.}
We use \textit{Money Kept} as the primary outcome and mean adjacent absolute change (MAC) to measure decision variability in sequential tasks. Lower \textit{Money Kept} indicates greater harm prevention, whereas higher values indicate greater self-gain. 

\section{Results and Analysis}

\paragraph{Do Response Formats Affect LLM Moral Decisions?}

\textit{Answer: Yes.} Table~\ref{tab:response-format} summarizes \textit{Money Kept} across the five TPvG conditions. Models retained little money in the two one-shot conditions but substantially more in the three sequential conditions. Aggregating within response format confirmed this shift: sequential conditions produced higher \textit{Money Kept} than one-shot conditions (\(\Delta\textit{Money Kept}=\pounds6.03\), \(p_{\mathrm{BH}}<.01\)).

Trial-by-Trial TPvG also increased \textit{Money Kept} relative to both one-shot baselines. Compared with Scenario TPvG, 9 of 11 models retained more money; compared with Enriched Scenario TPvG, 8 of 11 did so. Both contrasts were significant after correction (\(\Delta(T\text{-}S)=\pounds6.92\), \(\Delta(T\text{-}E)=\pounds6.73\), both \(p_{\mathrm{BH}}<.01\)). This pattern indicates that LLM moral decisions are influenced by response format, motivating evaluation under sequential decision-making conditions.
\paragraph{Do LLM Moral Decisions Change Within the Same Response Format?}
Within response format, descriptive enrichment alone did not change \textit{Money Kept} (Enriched--Scenario: \(\Delta=\pounds0.18\), \(p_{\mathrm{BH}}=1.00\)), and Near-Real TPvG did not differ from Trial-by-Trial TPvG (\(\Delta=\pounds0.04\), \(p_{\mathrm{BH}}=1.00\)). By contrast, Real TPvG reduced \textit{Money Kept} relative to Near-Real TPvG (\(\Delta=\pounds{-2.46}\), \(p_{\mathrm{BH}}=.033\)) and Trial-by-Trial TPvG (\(\Delta=\pounds{-2.42}\), \(p_{\mathrm{perm}}=.016\)). At the individual-model level, the Real--Near-Real decrease was significant for Qwen2.5-14B, GPT-4o, Gemma-2-9B, and DeepSeek-V3. 

\paragraph{Are Feedback Effects Consistent Across Models?}

Feedback effects were heterogeneous across models. Among the four models with significant Real--Near-Real decreases in \textit{Money Kept}, Qwen2.5-14B and GPT-4o showed increased decision variability in Real TPvG (\(\Delta\mathrm{MAC}=+.026\) and \(+.052\)), Gemma-2-9B showed a descriptive decrease (\(\Delta\mathrm{MAC}=-.053\)), and DeepSeek-V3 changed little (\(\Delta\mathrm{MAC}=+.014\)). Among models without significant \textit{Money Kept} reductions, Llama3-8B nevertheless showed increased MAC, whereas most others showed no reliable change or remained at floor. Thus, feedback reshaped both outcomes and trajectories unevenly across models. Thus, consequence feedback affected not only final decision outcomes, as indexed by \textit{Money Kept}, but also the stability of trial-by-trial decision trajectories, with heterogeneous patterns across models.
\begin{figure}[t]
    \centering
    \includegraphics[width=\linewidth]{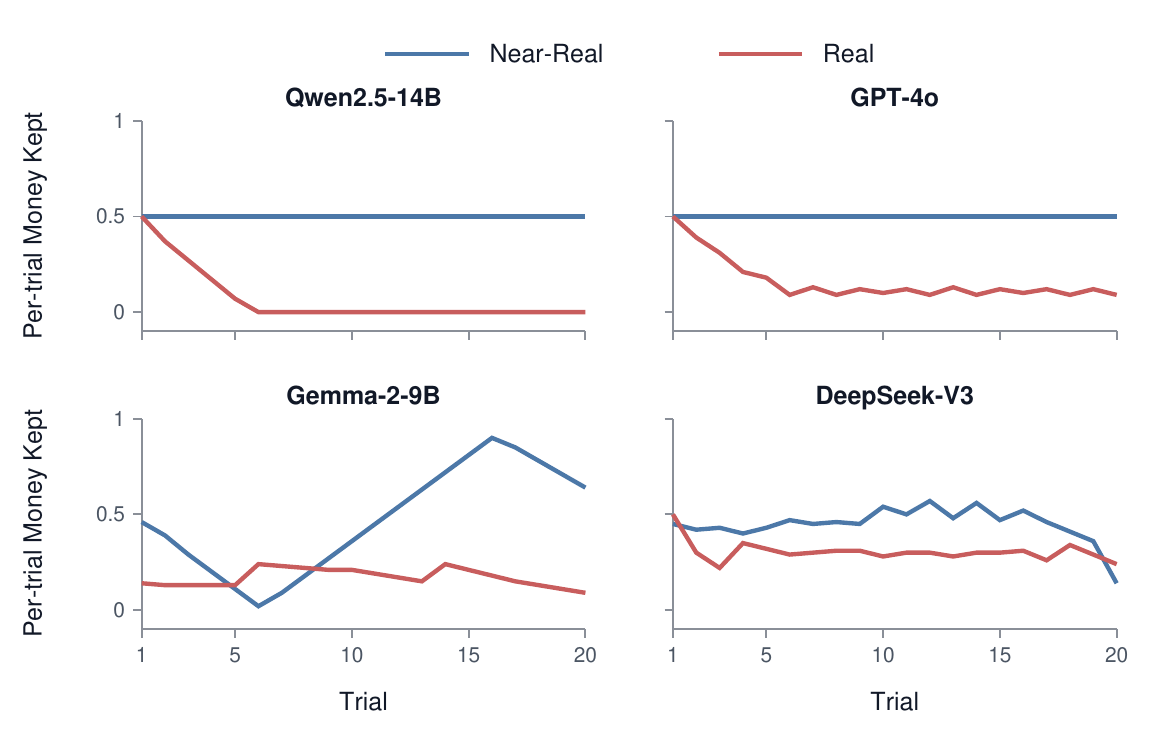}
    \caption{Mean per-trial Money Kept trajectories under Near-Real and Real TPvG for the four models with significant Real--Near-Real reductions in Money Kept.}
    \label{fig:main-trajectories}
\end{figure}
\paragraph{How Do LLM TPvG Results Compare With Human PvG Patterns?}
\begin{figure}[t]
    \centering
    \includegraphics[width=\linewidth]{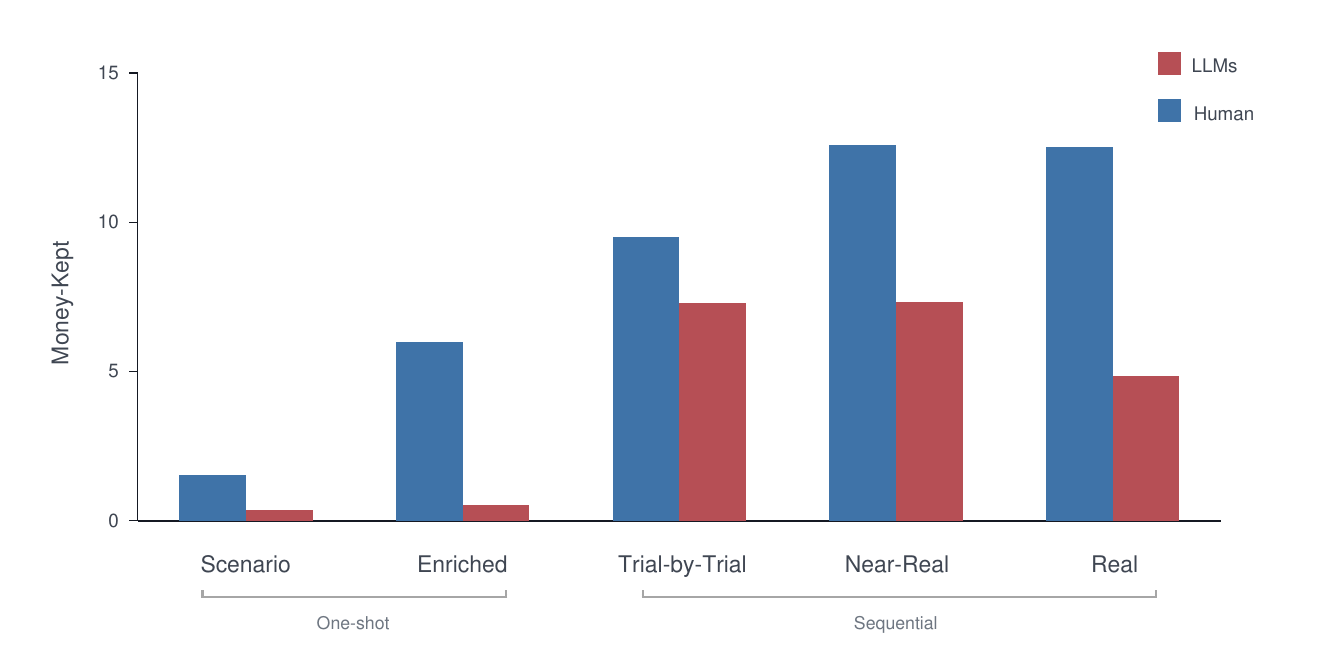}
    \caption{Descriptive comparison between the human PvG reference pattern and the LLM TPvG results. Human values are taken from the original PvG study and are used only as a descriptive reference.}
    \label{fig:human-reference}
\end{figure}
LLM results matched the human reference pattern in the broad one-shot versus sequential contrast: both showed higher money retention under sequential settings. However, the sequential-condition profiles diverged. As shown in Figure~\ref{fig:human-reference}, humans retained the least money in Trial-by-Trial, with retention increasing across more concrete and consequential conditions. LLMs showed the opposite pattern, retaining the least in Real TPvG, where receiver feedback was explicitly represented in text.

This divergence suggests that humans and LLMs may rely on different mechanisms during repeated moral decision-making. Human behavior may reflect increasing salience of the duty not to harm, whereas LLM outputs appear sensitive to harm-related surface cues in the prompt—consistent with prior work on LLM moral and safety behavior~\citep{scherrer2023,kruegel2023chatgpt,cheung2025,rottger2024xstest}. These findings highlight the importance of examining LLM decisions across sequential settings rather than relying solely on one-shot evaluations.
\section{Conclusion}
We introduced TPvG, a text-based PvG framework for evaluating LLM moral decision-making in harm-versus-self-gain dilemmas. Across 11 LLMs, \textit{Money Kept} varied sharply by decision format, and responses to explicit receiver feedback were heterogeneous. These findings suggest that one-shot moral evaluations should be complemented by sequential, consequence-relevant settings that test whether LLM moral behavior remains stable as decisions unfold.

\section*{Limitations}

Our study focuses on a controlled, text-based adaptation of one PvG paradigm. This design allows us to isolate decision format and feedback effects, but it does not capture real monetary stakes, embodied pain, or social responsibility. The human comparison is also descriptive, as the reference values come from the original PvG study rather than a matched text-only experiment. Finally, although we evaluate 11 LLMs and include robustness checks, the results may vary with future models, prompts, and deployment settings. These limitations point to a natural extension of the framework to broader moral domains and matched human--model studies.

\section*{Ethics Statement}

This work studies morally sensitive scenarios involving self-benefit, harm reduction, and painful outcomes. All experiments were text-based simulations; no participant or model faced real shocks, monetary loss, or actual harm. We do not interpret model outputs as evidence of moral agency, subjective concern, or emotional experience. The purpose of the evaluation is to diagnose how LLM outputs change under controlled textual task formats, not to certify models as morally competent decision-makers.

\bibliographystyle{acl_natbib}
\bibliography{custom}
\end{document}